\documentclass{article}

\usepackage{enumitem}
\usepackage[preprint]{neurips_2026}

\usepackage[utf8]{inputenc} 
\usepackage[T1]{fontenc}    
\usepackage{hyperref}       
\usepackage{url}            
\usepackage{booktabs}       
\usepackage{amsfonts}       
\usepackage{nicefrac}       
\usepackage{microtype}      
\usepackage{xcolor}         
\usepackage{amsmath, amssymb}
\usepackage{subcaption}
\usepackage{float}
\usepackage{multirow}
\usepackage{tikz}
\usetikzlibrary{arrows.meta,positioning}
\usepackage{graphicx}

\DeclareMathOperator*{\argmax}{arg\,max}

\usepackage{comment}

\usepackage{tcolorbox}        
\newcounter{formulation}
\renewcommand{\theformulation}{\arabic{formulation}}
\newtcolorbox{formulation}[2][]{colback=white, colframe=gray, boxrule=0.5mm, sharp corners, top=-2pt, left=2.5pt, right=2.5pt, title=Formulation \theformulation. #2, #1}

\newcommand{\CPCF}{\textsc{Cpcf}}
\newcommand{\OCEAN}{\textsc{Ocean}}
\newcommand{\MaxSAT}{\textsc{MaxSAT}}
\newcommand{\MACE}{\textsc{Mace}}

\title{Optimal Counterfactual Search in Tree Ensembles: \\ A Study Across Modeling and Solution Paradigms}

\author{%
  Awa Khouna \\
  Polytechnique Montréal\\
  \texttt{awa.khouna@polymtl.ca} \\
  \And   
  Youssouf Emine \\
  Ivey Business School\\
  \texttt{yemine@ivey.ca} \\
  \And 
  Julien Ferry \\
  Polytechnique Montréal\\
  \texttt{julien.ferry@polymtl.ca} \\
  \And 
  Thibaut Vidal \\
  Polytechnique Montréal\\
  \texttt{thibaut.vidal@polymtl.ca} \\
}

\begin{document}

\maketitle

\begin{abstract}
Trust in counterfactual explanations depends critically on whether their recommended changes are truly minimal: suboptimal explanations may vastly overshoot the actual changes needed to alter a decision, and heuristic errors can affect individuals unevenly, giving some users relevant recourse while assigning others unnecessarily costly recommendations. Consequently, we study the problem of computing \emph{optimal} counterfactual explanations for tree ensembles under plausibility and actionability constraints. This is a combinatorial problem: for a fixed model, counterfactual search boils down to selecting consistent branching decisions and threshold-defined regions under a distance objective. We exploit this structure through \CPCF, a constraint programming (CP) formulation in which numerical features are encoded as interval domains induced by split thresholds, while discrete features retain native finite-domain representations. This yields a compact finite-domain formulation that supports multiple distance objectives without continuous split-boundary search. We then place \CPCF{} in a broader comparison across mathematical programming paradigms: we extend a maximum Boolean satisfiability (MaxSAT) formulation, originally designed for hard-voting random forests, to soft-voting ensembles, and compare against the current state-of-the-art mixed-integer linear programming (MILP) optimal approach. Across ten datasets and three types of tree ensembles, we analyze scalability, anytime performance, and sensitivity to distance metrics. We observe that CP achieves the best overall performance. More importantly, our results identify regimes in which the specific strengths of each paradigm make it best suited: CP is most versatile overall, MaxSAT handles hard-voting ensembles particularly well, and MILP remains competitive in amortized inference settings with a moderate number of split levels.
\end{abstract}

\section{Introduction}

Counterfactual explanations have become a widely used form of local explanation for machine learning models~\citep{karimi2022survey,verma2024counterfactual}. Given an input instance, a counterfactual specifies how its features should change to obtain a desired prediction, thereby providing interpretable recourse in domains such as credit, healthcare, and public decision-making.
However, generating high-quality counterfactuals can be challenging for many models of interest: their decision boundaries may be non-convex and non-differentiable, their feature spaces can combine different feature types, and realistic explanations must satisfy plausibility and feasibility constraints~\citep{joshi2019realisticindividualrecourseactionable,poyiadzi2020face}. As a result, many counterfactual approaches rely on heuristics or generative models~\citep{vanlooveren2020interpretablecounterfactualexplanationsguided,nemirovsky2022countergan}, without guarantees on the quality of the returned solution.
This limitation has severe consequences. Since counterfactual explanations are often interpreted as minimal changes required to alter a decision, suboptimal outputs can overstate the required effort and suggest unnecessarily costly or impractical forms of recourse \citep{parmentier2021optimal}. Errors may also be unevenly distributed across the input space: some individuals may receive near-minimal recommendations, while others are assigned substantially more burdensome changes than necessary. This motivates the study of \emph{optimal} counterfactual explanations, which provably minimize a prescribed notion of change under prediction and feasibility constraints.

Tree ensembles are a central model class in this context. They remain among the strongest predictors for many tabular-data tasks that require explanations and recourse, yet their non-convex, non-differentiable decision functions pose substantial challenges for counterfactual generation. At the same time, their structure exposes a combinatorial problem that can be leveraged by optimization algorithms: for any fixed ensemble, counterfactual search reduces to selecting consistent split decisions and leaf regions.
Several heuristics exploiting this structure have been proposed. Early methods such as actionable feature tweaking and related recourse procedures search for favorable leaf regions or candidate modifications without fully solving the underlying optimization problem \citep{Tolomei_2017,cui2015,fernandez2020random}. Other approaches prioritize speed, diversity, or realism, often through approximate search or data-driven restrictions of the feasible region \citep{carreira2023LIRE,ZHANG2025128661}. Differentiable approximations of the decision boundaries of the ensembles have also been explored~\citep{lucic2022focus}. 
However, these heuristic algorithms regularly lead to vastly suboptimal explanations, e.g., by factors of over $10 \times$ as seen in \citep{parmentier2021optimal,khouna2026counterfactualmaps}. To yield explanations with guarantees, recent work has explored mathematical programming formulations, based on mixed-integer linear programming (MILP)~\citep{parmentier2021optimal}, satisfiability-based methods~\citep{karimi2020modelagnostic}, and weighted maximum Boolean satisfiability (MaxSAT)~\citep{inbookMaxSAT}. These approaches differ in how they represent and explore the search space, and on which mathematical programming paradigm they are grounded to prune the search space and ensure optimality.

\begin{figure}[t]
\centering
\resizebox{\linewidth}{!}{%
\begin{tikzpicture}[
    >=Latex,
    node distance=6mm and 13mm,
    decision/.style={
        rectangle,
        rounded corners=2pt,
        draw=black,
        thick,
        align=center,
        minimum width=30mm,
        minimum height=8mm,
        font=\footnotesize,
        fill=gray!8
    },
    leafC/.style={
        rectangle,
        rounded corners=2pt,
        draw=orange!70!black,
        thick,
        align=center,
        minimum width=15mm,
        minimum height=7mm,
        inner sep=1.5pt,
        font=\footnotesize\bfseries,
        fill=orange!18
    },
    leafM/.style={
        rectangle,
        rounded corners=2pt,
        draw=green!50!black,
        thick,
        align=center,
        minimum width=15mm,
        minimum height=7mm,
        inner sep=1.5pt,
        font=\footnotesize\bfseries,
        fill=green!18
    },
    leafX/.style={
        rectangle,
        rounded corners=2pt,
        draw=violet!70!black,
        thick,
        align=center,
        minimum width=15mm,
        minimum height=7mm,
        inner sep=1.5pt,
        font=\footnotesize\bfseries,
        fill=violet!18
    },
    edge/.style={draw, thick, -{Latex[length=1.8mm]}}
]

\node[decision] (root) {Voting\\{\footnotesize 100.0\%}};

\node[decision, right=15mm of root, yshift=10mm] (metric) {Performance metric \\{\footnotesize 67.2\%}};
\node[decision, right=15mm of root, yshift=-10mm] (ord) {\#Ordinal features \\{\footnotesize 32.8\%}};

\node[decision, right=20mm of metric, yshift=0mm] (splits) {Mean \#split levels\\per feature \\{\footnotesize 34.3\%}};
\node[decision, right=20mm of ord, yshift=3mm] (estim) {\#Estimators\\{\footnotesize 6.8\%}};

\node[leafM, right=24mm of splits, yshift=4mm] (mip) {MILP\\{\footnotesize 13.5\%}};
\node[leafC, right=24mm of splits, yshift=-9mm] (cpcf) {CPCF\\{\footnotesize 58.8\%}};
\node[leafX, right=24mm of splits, yshift=-21mm] (maxsat) {MAXSAT\\{\footnotesize 27.7\%}};

\coordinate (cpcf_in_top) at ([yshift=2mm]cpcf.west);
\coordinate (cpcf_in_bot) at ([yshift=-2mm]cpcf.west);
\coordinate (maxsat_in_bot) at ([yshift=-1mm]maxsat.west);

\path[use as bounding box]
    ([yshift=2mm]root.north west)
    rectangle
    ([yshift=-2mm]maxsat.south east);

\draw[edge] (root) -- node[above, sloped, font=\footnotesize] {Soft} (metric);
\draw[edge] (root) -- node[below, sloped, font=\footnotesize] {Hard} (ord);

\draw[edge] (metric) to [bend right=5] node[below, pos=0.1, sloped, font=\footnotesize] {total time} (cpcf.west);
\draw[edge] (metric) --  node[above, sloped, font=\footnotesize] {only solver time} (splits.west);

\draw[edge] (ord) to [bend right=8] node[above, pos=0.1, sloped, font=\footnotesize] {$\geq 2$} (maxsat_in_bot);
\draw[edge] (ord) -- node[above, sloped, font=\footnotesize] {$< 2$} (estim.west);

\draw[edge] (splits.east) to[bend left=12] node[above, sloped, font=\footnotesize] {$\leq 7.42$} (mip.west);
\draw[edge] (splits.east) to[bend right=6]  node[above, sloped, font=\footnotesize] {$> 7.42$} (cpcf_in_top);

\draw[edge] (estim.east)  to[bend left=10] node[below, sloped, font=\footnotesize] {$\leq 100$} ([yshift=1mm]maxsat.west);
\draw[edge] (estim.east)  to[bend right=10]  node[pos=0.3, above, sloped, font=\footnotesize] {$> 100$} (cpcf_in_bot);

\end{tikzpicture}%
}
\vspace*{0.1cm}
\caption{Decision diagram summarizing the best-performing optimal counterfactual formulation for tree ensembles, for each setting (see Appendix~\ref{app:decision-diagram} for construction details). Percentages indicate the share of experimental configurations reaching each node. The proposed \CPCF{} performs best overall, \MaxSAT{}~\citet{inbookMaxSAT} excels on moderate-size hard-voting random forests or datasets with many ordinal features, while MILP~\citep{parmentier2021optimal} is competitive in soft-voting settings with moderately many split levels per feature, especially if only solver time matters (e.g., if optimization model-building time is amortized over many queries).} \label{fig:decision-diagram}
\end{figure}
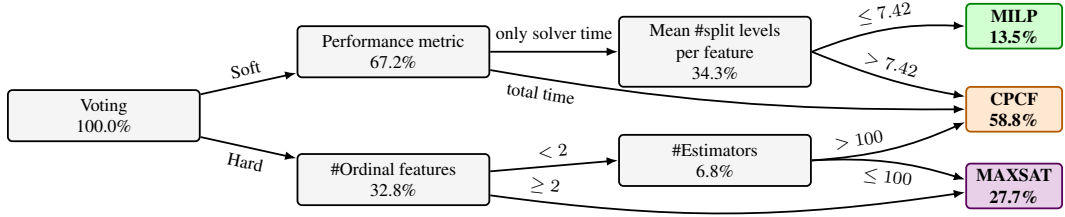

In this work, we first introduce \CPCF, a constraint programming formulation for optimal counterfactual search in tree ensembles. Our starting point is that the finite-domain primitives of CP align naturally with the split-induced regions of tree ensembles: numerical features can be represented through threshold intervals, discrete features retain their finite-domain structure, and tree paths are enforced through logical constraints. This yields a compact formulation that directly exploits the problem's combinatorial structure and supports multiple distance objectives, as well as plausibility and actionability requirements. Compared to MILP formulations, \CPCF\ avoids reasoning over continuous split boundaries during search and does not require linearization; compared to Boolean encodings, it retains a compact representation of numerical features. We then place \CPCF\ in a broader empirical comparison of exact mathematical programming paradigms for counterfactual search. Across ten datasets, multiple ensemble types, and several distance objectives, \CPCF\ achieves the best overall performance. More importantly, our analysis shows that no single paradigm dominates across all regimes: CP is the most versatile overall, MaxSAT is particularly effective for hard-voting random forests, and MILP is competitive when model-building time can be amortized in moderate split-level regimes. This provides practical guidance for selecting an exact counterfactual solver, summarized by the decision diagram in Figure~\ref{fig:decision-diagram}, which relates the preferred formulation to ensemble characteristics, dataset structure, and the performance metric of interest. 
Our main contributions are as follows:
\begin{enumerate}[leftmargin=!, labelindent=5pt, topsep=1pt, itemsep=1pt, parsep=0pt]
    \item We introduce \CPCF, the first constraint programming formulation for optimal counterfactual explanations in tree ensembles. It provides a compact finite-domain model of counterfactual search and supports common tree ensembles, heterogeneous feature types, distance objectives, plausibility and actionability requirements.

    \item We provide a unified implementation of exact mathematical programming approaches for this problem, spanning CP, MILP, SAT, and MaxSAT. As part of this framework, we extend the existing MaxSAT formulation of~\citet{inbookMaxSAT} beyond hard-voting random forests to support soft-voting ensembles and boosted trees. A user-friendly library incorporating all formulations will be released upon acceptance, under an MIT license.

    \item We conduct a large-scale empirical comparison across ten datasets with varied characteristics, multiple ensemble types, and several distance objectives. Beyond aggregate performance, our analysis characterizes when each paradigm is preferable, providing practical guidance for selecting an exact counterfactual solver.
\end{enumerate}

\section{Background}
\label{sec:Background}

Let \(\mathcal{T}\) be a tree ensemble (e.g., random forest or XGBoost), with classification function \(h_\mathcal{T} : \mathcal{X} \mapsto \mathcal{Y}\). For any given input sample \(\mathbf{x} \in \mathcal{X}\), its predicted label is the class with maximum confidence score:
\begin{equation*}
h_\mathcal{T}(\mathbf{x}) = \argmax_{y \in \mathcal{Y}} s_y(\mathbf{x})
\quad \text{where} \quad
s_y(\mathbf{x}) = b_y + \sum_{t \in \mathcal{T}} w_t s_{t,y}(\mathbf{x}),
\end{equation*}
where $b_y$ is a constant bias for class $y$, \(w_t\) denotes the weight associated with tree \(t \in \mathcal{T}\), and \(s_{t,y}(\mathbf{x})\) denotes the score assigned by tree \(t\) to class \(y\). 

Two \emph{voting schemes} are commonly used. Hard voting occurs when each tree casts a vote for a single class, i.e., \(w_t = 1\) and \(s_{t,y}(\mathbf{x}) \in \{0,1\}\), whereas soft voting generalizes this setting by allowing each tree to contribute class-specific confidence scores, with \(w_t \in [0,1]\) and \(s_{t,y}(\mathbf{x}) \in \mathbb{R}\)
, for each tree \(t \in \mathcal{T}\) and class \(y \in \mathcal{Y}\).

Let \(\hat{\mathbf{x}} \in \mathcal{X}\) denote a \emph{query} sample and let \(y^\star \in \mathcal{Y}\) denote a desired \emph{target class}, such that $h_\mathcal{T}(\hat{\mathbf{x}}) \neq y^\star$. An \emph{optimal counterfactual explanation} for \(\hat{\mathbf{x}}\) is a sample \(\mathbf{x} \in \mathcal{X}\) that is classified as the target class \(y^\star\), lies as close as possible to \(\hat{\mathbf{x}}\) under a prescribed cost function, and satisfies optional plausibility and actionability constraints. We therefore consider the optimization problem:
\begin{equation}
\label{eq:problem}
\min_{\mathbf{x} \in \mathcal{X}} \mathrm{cost}(\mathbf{x},\hat{\mathbf{x}})
\quad \text{s.t.} \quad
s_{y^\star}(\mathbf{x}) \geq s_y(\mathbf{x}) + \varepsilon_c,
\quad \mathbf{x} \in \mathcal{X}_{\mathrm{plausible}} \cap \mathcal{X}_{\mathrm{actionable}},
\qquad \forall y \in \mathcal{Y}\setminus\{y^\star\},
\end{equation}
where \(\mathrm{cost}(\mathbf{x},\hat{\mathbf{x}})\) measures the cost of changing the query instance into the counterfactual, and \(\varepsilon_c > 0\) is a small margin used to enforce a strict target prediction. The set \(\mathcal{X}_{\mathrm{plausible}}\) restricts counterfactuals to regions well supported by the empirical data distribution, excluding low-density or out-of-distribution instances, while \(\mathcal{X}_{\mathrm{actionable}}\) encodes the values attainable from \(\hat{\mathbf{x}}\) under structural, immutability, and monotonicity constraints.
This formulation is shared by all methods considered in this paper. The differences between \CPCF\ and the baselines therefore do not lie in the high-level optimization problem, but rather in how the feasible set, tree structure, and prediction constraints are encoded within different mathematical programming paradigms.

Several mathematical programming formulations have been proposed to model and solve Problem~\eqref{eq:problem}. First, the seminal work of \citet{karimi2020modelagnostic}, known as \MACE, maps the nearest-counterfactual problem into a sequence of satisfiability problems, encoding the predictive model, the distance function, and feasibility requirements as logical formulae. Subsequently, \OCEAN~\citep{parmentier2021optimal} proposed a MILP formulation for tree ensembles, in which feature changes, tree paths, and class scores are represented through linear constraints and both continuous and binary decision variables. More recently, \citet{inbookMaxSAT} leveraged a partial weighted Maximum Satisfiability (\MaxSAT) formulation, where validity and feasibility conditions are encoded as hard clauses, while the counterfactual cost is represented through weighted soft clauses. This formulation is restricted to hard-voting ensembles, since incorporating numerical weights and confidence scores requires pseudo-Boolean encodings and pairwise class-score comparisons, which can be computationally costly in a purely propositional encoding. For comparison purposes, we extend it to handle the general case of soft-voting. To this end, we encode each pairwise target-class constraint as a hard pseudo-Boolean inequality over leaf-selection variables, and translate these inequalities into weighted CNF using a standard pseudo-Boolean encoding \citep{pseudobool, pblib.sat2015}. This extends the original \MaxSAT\ formulation beyond hard-voting while preserving a purely propositional optimization model (see details in Appendix~\ref{app:maxsat-soft-voting}).

Our proposed \CPCF\ formulation instead leverages constraint programming (CP)~\citep{rossi2006handbook}, a mathematical programming paradigm that models combinatorial problems through decision variables, finite domains, and constraints. A key strength of CP lies in its ability to exploit global constraints, which provide compact, high-level representations of structured subproblems and dedicated filtering algorithms. These algorithms propagate constraints by removing inconsistent values from variable domains, thereby pruning large portions of the search space before and during the search. In contrast to MILP formulations, CP does not require all constraints to be linearized, and can instead preserve and exploit the discrete and logical structure of the counterfactual search problem in tree ensembles.
Another distinctive feature of \CPCF\ is that numerical and ordinal features are represented through interval domains induced directly by the split thresholds appearing in the ensemble.

\section{Methodology}\label{sec:mathematical_programming_formulations}

We now describe how \CPCF\ encodes and solves Problem~\eqref{eq:problem}.
For each tree $t \in \mathcal{T}$, we use $\mathcal{L}_t$ to denote its set of leaves. Each leaf $\ell \in \mathcal{L}_t$ is associated to a confidence score $p_{t,\ell,y}$ for each possible class $y \in \mathcal{Y}$, such that $s_{t,y}(\mathbf{x})=p_{t,\ell,y}$ if example $\mathbf{x}$ falls into leaf $\ell$ of tree $t$.

\paragraph{Modeling the counterfactual features.}
For each feature \(f \in \{1,\dots,d\}\), we define one decision variable encoding the counterfactual's value, according to the feature type:
\begin{itemize}[leftmargin=!, labelindent=5pt, topsep=1pt, itemsep=1pt, parsep=0pt]
\item Each \emph{binary feature} is modeled through a binary variable \(x_f \in \{0,1\}\).
    \item Each \emph{categorical feature} is represented through a one-hot encoding group \(G_k \subseteq \{1,\dots,d\}\) of binary variables \(x_f \in \{0,1\}\). To ensure that exactly one category is selected in each group, we enforce \(\sum_{f \in G_k} x_f = 1\) via a global \texttt{ExactlyOne} constraint.
    \item \emph{Numerical features} are modeled through interval encoding. Let \(A_f = \{\tau_{f,1},\dots,\tau_{f,k_f}\}\) be the sorted set of split thresholds appearing in \(\mathcal{T}\) for feature \(f\). Together with global lower and upper bounds \(\mathrm{lb}(f)\) and \(\mathrm{ub}(f)\), these thresholds define the ordered partition
    \[
        \mathcal{I}_f = \big(\mathrm{lb}(f),\tau_{f,1},\dots,\tau_{f,k_f},\mathrm{ub}(f)\big).
    \]
    Instead of reasoning on the feature values directly, \CPCF{} introduces an interval-index integer variable \(x_f \in \{0,\dots,k_f\}\) for each feature  \(f\), where \(x_f = m\) indicates that its value belongs to the interval \((\mathcal{I}_f[m],\,\mathcal{I}_f[m+1]]\)\footnote{For XGBoost ensembles, we use \([\mathcal{I}_f[m],\,\mathcal{I}_f[m+1))\).}. Note that this discretization remains exact with respect to the ensemble: two values in the same interval are indistinguishable to the model, as they induce identical tree routing decisions.
\end{itemize}

\begingroup
\setlength{\abovedisplayskip}{3pt}
\setlength{\belowdisplayskip}{3pt}
\setlength{\abovedisplayshortskip}{2pt}
\setlength{\belowdisplayshortskip}{2pt}

\paragraph{Modeling the counterfactual paths.}
We use a binary variable \(z_{t,\ell} \in \{0,1\}\) to indicate whether the counterfactual \(\mathbf{x}\) falls into leaf \(\ell \in \mathcal{L}_t\) of tree \(t \in \mathcal{T}\). Since the counterfactual falls into exactly one leaf of each tree, we leverage global \texttt{ExactlyOne} constraints to impose
\begin{align}
\label{eq:leaf}
\sum_{\ell \in \mathcal{L}_t} z_{t,\ell} = 1 \qquad \forall t \in \mathcal{T}.
\end{align}
Each leaf corresponds to a conjunction of split conditions along its root-to-leaf path. We therefore enforce consistency between leaf assignments and counterfactual feature values:
\begin{itemize}[leftmargin=!, labelindent=5pt, topsep=1pt, itemsep=1pt, parsep=0pt]
    \item For \emph{numerical features}, split conditions are enforced through the corresponding interval index. If the path to leaf \(\ell\) in tree \(t\) contains the condition \(x_f \le \tau\), then
    \begin{equation}
    \label{eq:cp_num_left}
    z_{t,\ell}=1 \;\Rightarrow\; x_f \le m_f(\tau),
    \end{equation}
    where \(m_f(\tau)\) is the index of threshold \(\tau\) in the ordered partition \(\mathcal{I}_f\). Likewise, if the path contains \(x_f > \tau\), then
    \begin{equation}
    \label{eq:cp_num_right}
    z_{t,\ell}=1 \;\Rightarrow\; x_f \ge m_f(\tau)+1.
    \end{equation}

    \item For \emph{binary and one-hot encoded categorical features}, path conditions are enforced directly on the corresponding binary variable. If the path to leaf \(\ell\) requires feature \(f\) to take value \(v \in \{0,1\}\), then
    \begin{equation}
    \label{eq:cp_binary_path}
    z_{t,\ell}=1 \;\Rightarrow\; x_f = v.
    \end{equation}
\end{itemize}
Implications~\eqref{eq:cp_num_left} to \eqref{eq:cp_binary_path} are implemented using efficient reification constraints. Since the split conditions within each tree induce a partition of the feature space, these implications implicitly enforce the leaf-selection constraints in~\eqref{eq:leaf}. However, explicitly imposing them through global constraints yields stronger and faster propagation.

\endgroup

\paragraph{Enforcing target class prediction.} The class score for each class \(y \in \mathcal{Y}\) is computed as
\begin{equation}
\label{eq:score}
s_y = b_y + \sum_{t \in \mathcal{T}} \sum_{\ell \in \mathcal{L}_t} w_t p_{t,\ell,y} \, z_{t,\ell},
\end{equation}
where \(w_t\) is the weight of tree \(t\), \(b_y\) is an optional class-dependent base score, and \(p_{t,\ell,y}\) denotes the contribution of leaf \(\ell\) in tree \(t\) to class \(y\), so that \(s_{t,y}(\mathbf{x}) = \sum_{\ell \in \mathcal{L}_t} p_{t,\ell,y} z_{t,\ell}\).
We then ensure that the counterfactual is predicted as the target class \(y^\star\) by imposing
\begin{align}
\label{eq:target_class}
s_{y^\star} \geq s_y + \varepsilon_c
\qquad \forall y \in \mathcal{Y}\setminus\{y^\star\}.
\end{align}

\paragraph{Objective value.}
As in Problem~\eqref{eq:problem}, the objective is to minimize the distance, or cost, between the query instance \(\hat{\mathbf{x}}\) and the counterfactual \(\mathbf{x}\). In \CPCF, this objective is represented featurewise, with per-feature actionability costs. For separable distances, \CPCF{} optimizes
\begin{equation}
\label{eq:cp_sep_obj}
\min_{\mathbf{x},\mathbf{z}} \mathrm{cost}(\mathbf{x},\hat{\mathbf{x}})
=
\sum_{f=1}^d \alpha_f \delta_f(x_f,\hat{x}_f),
\end{equation}
where \(\alpha_f\) is a user-specified feature-specific cost coefficient.
For binary and categorical features, \(\delta_f(x_f,\hat{x}_f)\) is defined directly from the change in feature value. For numerical features, the distance depends on the selected interval. For instance, using the \(L_p\)-type separable cost with \(p \geq 0\), we define
\begin{equation}
\label{eq:cp_delta_num}
\delta_f(m,\hat{x}_f)=
\begin{cases}
0, & \text{if } \hat{x}_f \in [\mathcal{I}_f[m],\,\mathcal{I}_f[m+1]),\\[1mm]
(\mathcal{I}_f[m]-\hat{x}_f)^p, & \text{if } \hat{x}_f < \mathcal{I}_f[m],\\[1mm]
(\hat{x}_f-\mathcal{I}_f[m+1])^p, & \text{if } \hat{x}_f \ge \mathcal{I}_f[m+1],
\end{cases}
\end{equation}
which corresponds to the minimum displacement required to move the query value into the selected interval. In practice, the cost associated with each interval is retrieved from its index, leveraging efficient \texttt{Element} constraints. More elaborate objectives, such as nonlinear actionability costs or direction-dependent costs, can be incorporated by modifying the per-interval costs computed in~\eqref{eq:cp_delta_num}.

\paragraph{Actionability and plausibility.}
The plausibility and actionability restrictions appearing in Problem~\eqref{eq:problem} can be imposed directly on the feature variables. In particular, immutable features can be fixed to their original values, directional constraints can enforce monotonic changes, and admissible-domain restrictions can be added independently of the ensemble structure. This modularity is one of the practical advantages of the CP formulation, whose large and generic catalog of constraints allows additional desiderata to be expressed efficiently. Furthermore, similar to~\citet{parmentier2021optimal}, plausibility can be enforced in \CPCF{} through a user-provided isolation forest~\citep{liu2008isolation}, a popular outlier-detection model that computes anomaly scores from the average path length of an example across a set of randomly built trees. As detailed in Appendix~\ref{app:isolation-details}, the path of the counterfactual through each isolation tree is then modeled using constraints similar to Constraints~\eqref{eq:leaf}--\eqref{eq:cp_binary_path}, and a hard constraint is imposed on the resulting anomaly score.

\section{Numerical Experiments}\label{sec:experimental-results}

This section evaluates \CPCF{} alongside existing mathematical programming formulations on benchmark cases spanning heterogeneous feature types, ensemble types, and controlled variations in ensemble complexity. We assess the overall efficiency of \CPCF{} and characterize how each paradigm behaves across tree depth, ensemble size, cost function, and plausibility constraints.

\subsection{Experimental Setup}

\textbf{Baselines.}
We evaluate \CPCF{} alongside three methods spanning the different mathematical programming paradigms for optimal counterfactual generation in tree-based models: \OCEAN, current state-of-the-art MILP formulation~\citep{parmentier2021optimal}; \MACE, a satisfiability-based approach~\citep{karimi2020modelagnostic}; and the \MaxSAT~\citep{inbookMaxSAT} formulation. For \OCEAN, we use the original formulation of~\citet{parmentier2021optimal}, with minor solver-parameter adjustments to preserve fidelity to the underlying \texttt{scikit-learn} random forest implementation. For \MACE, we set \(\varepsilon = 10^{-3}\).

\CPCF{} is solved using the \texttt{OR-Tools CP-SAT} solver v9.14.6206~\citep{cpsatlp}, \OCEAN{} using \texttt{Gurobi} v12.0, \MaxSAT{} using the \texttt{CaDiCaL} SAT solver~\citep{BiereFallerFazekasFleuryFroleyks}, and \MACE{} using \texttt{Z3}~\citep{Z3_solver}, all through their Python bindings.
Further details on numerical precision and method-specific implementation choices are provided in Appendix~\ref{app:backend-details}.

\textbf{Datasets.}
We use ten tabular classification datasets with heterogeneous feature types, summarized in Table~\ref{tab:datasets}. For AD, CC, CP, GC, ON, PH, SP, and ST, we follow the benchmark introduced by \citet{parmentier2021optimal}, in order to maintain direct comparability with prior exact approaches for tree-ensemble counterfactual explanations. We do not impose dataset-specific actionability restrictions, and complement this benchmark with two UCI datasets, namely BC \citep{breast_cancer_wisconsin_original_15} and SE \citep{seeds_236}. BC adds a small-scale binary classification task with predominantly ordinal features, while SE introduces a multiclass setting that is absent from the original benchmark. For BC, we remove incomplete observations, retaining 683 complete samples.

\begin{table}[t]
\centering
\small
\caption{Summary of the datasets used in our experiments. Columns \(N\), \(O\), \(C\), and \(B\) report the number of numerical, ordinal, categorical, and binary features, respectively. Categorical features are one-hot encoded during preprocessing, resulting in a total of \#$\mathcal{X}$ (OHE) features and \#$\mathcal{Y}$ classes.}
\label{tab:datasets}
\begin{tabular}{lrr|rrrr|r|r}
\toprule
Dataset & \# samples & \# features & \(N\) & \(O\) & \(C\) & \(B\) & \#$\mathcal{X}$ (OHE) & \#$\mathcal{Y}$ \\
\midrule
AD: Adult                   & 45,222 & 11 &  2 &  3 & 4 &  2 & 41 & 2 \\
BC: Breast Cancer Wisconsin &    683 &  9 &  0 &  9 & 0 &  0 &  9 & 2 \\
CP: COMPAS                  &  5,278 &  5 &  0 &  2 & 0 &  3 &  5 & 2 \\
CC: Credit Card             & 29,623 & 14 &  0 & 11 & 0 &  3 & 14 & 2 \\
GC: German Credit           &  1,000 &  9 &  1 &  5 & 3 &  0 & 19 & 2 \\
ON: Online News Popularity  & 39,644 & 47 & 37 &  6 & 2 &  2 & 59 & 2 \\
PH: Phishing                & 11,055 & 30 &  0 &  8 & 0 & 22 & 30 & 2 \\
SE: Seeds                   &    210 &  7 &  7 &  0 & 0 &  0 &  7 & 3 \\
SP: Spambase                &  4,601 & 57 & 57 &  0 & 0 &  0 & 57 & 2 \\
ST: Students Performance    &    395 & 30 &  0 & 13 & 4 & 13 & 43 & 2 \\
\bottomrule
\end{tabular}
\end{table}

\textbf{Models, configurations, and compute budget.}
For each dataset, we train tree ensemble classifiers and evaluate all counterfactual methods under a common protocol. We consider three ensemble types: hard-voting and soft-voting random forests~\citep{RF}, as implemented in the \texttt{scikit-learn} library~\citep{scikit-learn}, and XGBoost classifiers~\citep{chen2016xgboost} from the eponymous library, which inherently use soft-voting. Since \MACE{} does not support boosted ensembles, it is not benchmarked on XGBoost. Unless stated otherwise, the default configuration uses \(100\) estimators with maximum depth \(5\).
To isolate sources of combinatorial growth, we use two scaling protocols: depth scaling, with \(100\) estimators and maximum depth in \(\{3,4,5,6,7,8\}\), and ensemble scaling, with maximum depth \(5\) and number of estimators in \(\{10,20,50,100,200,500\}\). For each dataset and configuration, we sample \(50\) query instances and repeat experiments over five random seeds, yielding \(250\) counterfactual computations. All experiments run on homogeneous cluster nodes equipped with AMD EPYC 9654 (Zen 4) @ 2.40GHz CPUs, 64\,GB RAM, and 8 threads per run, with a \(900\)-second time limit per counterfactual search. The source code required to reproduce our experiments will be publicly released upon acceptance.

\textbf{Evaluation metrics.}
We focus on metrics that distinguish exact counterfactual methods. 
Since all methods solve the same optimization problem under identical prediction and feasibility constraints, standard counterfactual-quality measures such as validity, sparsity, and final distance are not informative for comparison: once optimality is certified, validity is guaranteed and the optimal objective value is the same across formulations. The central question is therefore computational: how quickly each method finds optimal counterfactuals and how they scale with the ensemble's characteristics.
We report final performance using median total generation time and optimality status. Total generation time includes both model-building time and solver time, where model-building time corresponds to instantiating the optimization model from the trained ensemble. This distinction matters because encoding overhead can differ substantially across paradigms; it also differs from prior work, which typically reports solver time only~\citep{parmentier2021optimal,inbookMaxSAT}. We also evaluate anytime performance by tracking the best incumbent objective value over time, thereby measuring solution quality before optimality is certified. Note that only \OCEAN{} and \CPCF{} support such anytime behavior. Finally, although all main experiments use the \(L_1\)-norm, the choice of recourse cost can substantially affect the structure of the optimization problem. We therefore assess cost sensitivity by comparing the two formulations that support all considered objectives, \CPCF{} and \OCEAN{}, under \(L_0\), \(L_1\), and \(L_2\)-type costs. We exclude \MaxSAT{} and \MACE{} from this analysis because they did not solve enough instances within the time limit to support a reliable comparison, and because \MaxSAT{} does not support the \(L_2\) objective. Further details regarding the experimental setup are provided in Appendix~\ref{app:experimental-details}.

\subsection{Results}
We organize the empirical analysis around five findings, covering default benchmark performance, anytime behavior, scaling with ensemble complexity, sensitivity to the objective, and the effect of plausibility constraints.

\begin{figure}[t]
    \centering
    \includegraphics[width=0.98\linewidth]{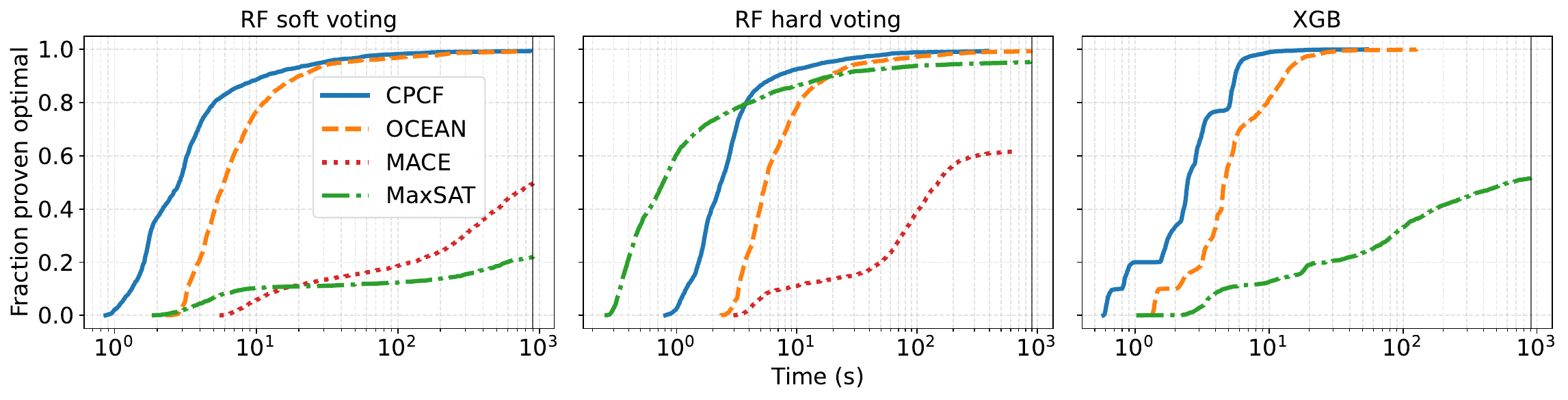}
    \caption{Cactus plots for the default ensemble configuration, aggregated over all ten datasets. Each curve reports the fraction of queries solved (out of a total of \(2{,}500\)) to proven optimality (higher is better) as a function of total time, including model construction.}
    \label{fig:default-cactus}
\end{figure}

\textbf{Result 1. The best-performing formulation depends on the setup.}
Figure~\ref{fig:default-cactus} shows the fraction of counterfactuals proven optimal as a function of total running time, aggregated over datasets and random seeds, for hard-voting random forests, soft-voting random forests, and XGBoost. The relative performance of the different methods is strongly regime-dependent, as summarized in Figure~\ref{fig:decision-diagram}. For hard-voting random forests, \MaxSAT{} is most effective, solving the largest fraction of instances the fastest, consistent with the target-class condition remaining close to a Boolean majority constraint. In soft-voting setups, however, it is no longer efficient, as it requires substantially larger Boolean encodings to model numerical confidence scores. In these settings, \CPCF{} performs best, proving optimality for almost all instances while achieving the fastest certification times. When model-build time is excluded, visible as the x-axis offset before search begins, \OCEAN{} and \CPCF{} exhibit very similar solver-time performance. \MACE{} is markedly less competitive in all tested configurations (and does not support XGBoost ensembles).

\begin{figure}[t]
    \centering
    \includegraphics[width=0.98\linewidth]{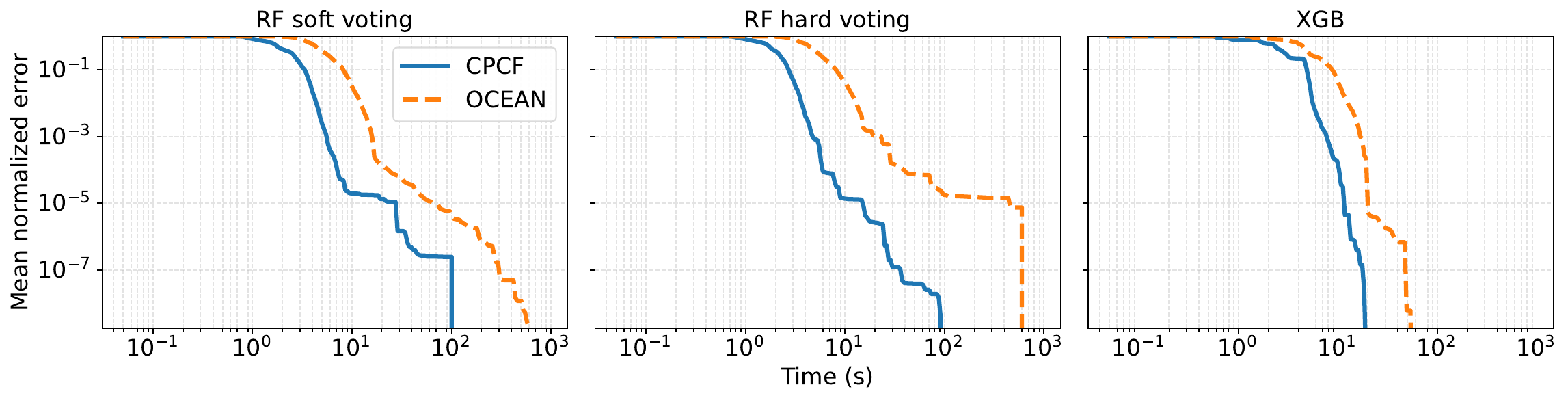}
    \caption{Anytime solution quality for \CPCF\ and \OCEAN\ for the default ensemble configuration, aggregated over all ten datasets and restricted to queries solved to optimality by both methods (which covers more than 99\% of the queries for all three ensemble types). The curve reports the mean normalized cost of the best incumbent objective; lower is better.}
    \label{fig:anytime}
\end{figure}

\textbf{Result 2. \CPCF\ reaches high-quality incumbents earlier than \OCEAN.}
Beyond certifying optimality, mathematical programming formulations such as CP and MILP can also be valuable because they provide anytime incumbent counterfactuals, together with lower bounds on the optimal cost, yielding a certifiable approximation gap for the current best solution. Figure~\ref{fig:anytime} compares \CPCF{} and \OCEAN{}, the only benchmarked methods supporting anytime results, through the best incumbent objective found over time. Across all three regimes, \CPCF{} improves the incumbent more rapidly in the early search phase. The difference is most pronounced for random forests, where the normalized counterfactual cost, with \(0\) denoting the optimal cost, decreases earlier and more sharply, indicating that \CPCF{} reaches near-optimal counterfactuals sooner. The same pattern remains visible for XGBoost, although with a smaller gap than for soft-voting random forests. Thus, when CP and MILP are the relevant broadly applicable paradigms, \CPCF{} is preferable not only because it certifies optimality faster, but also because it provides useful explanations earlier during the search.

\begin{figure}[t]
    \centering
    \begin{subfigure}{1.0\linewidth}
        \centering
        \includegraphics[width=\linewidth]{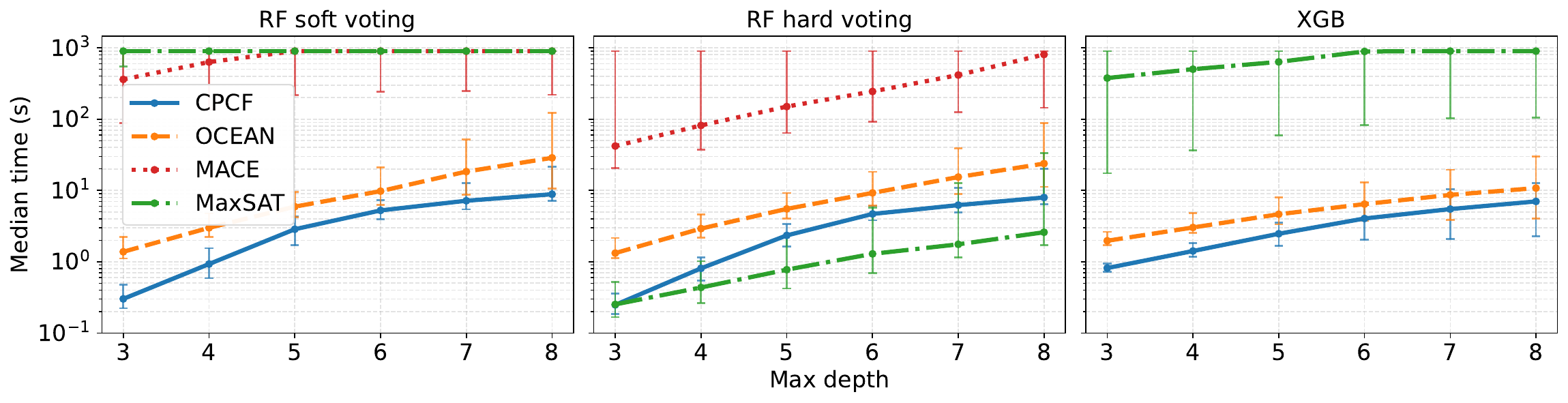}
        \caption{Scaling with the estimators' depth.}
        \label{fig:depth-scaling}
    \end{subfigure}
    \hfill
    \begin{subfigure}{1.0\linewidth}
        \centering
        \includegraphics[width=\linewidth]{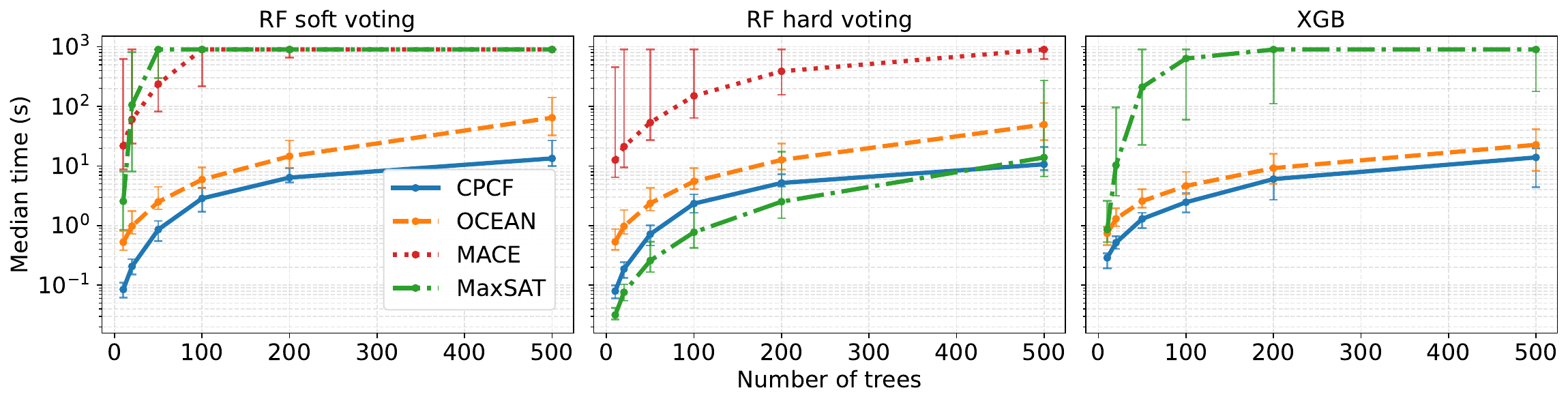}
        \caption{Scaling with the number of estimators.}
        \label{fig:estimator-scaling}
    \end{subfigure}
    \caption{Median total time to optimality (error bars indicate the first and third quartiles) with \(100\) trees of varying depth, and with varying numbers of trees of maximum depth $5$. Both panels aggregate all ten datasets. Missing or non-optimal runs are counted at the \(900\)-second time limit.}
    \label{fig:scaling}
\end{figure}

\textbf{Result 3. Solution methodologies differ in their scalability.}
We now study the impact of two characteristics of the ensembles that significantly impact the size of the search space: the number of trees and their depth. Figure~\ref{fig:scaling} reports the median total solution time as these parameters vary. For soft-voting random forests and XGBoost ensembles, \CPCF{} scales substantially better than \OCEAN{} across both the number of trees and their depth, while both remain far more efficient than the Boolean baselines. For hard-voting random forests, \MaxSAT{} retains the best performance across much of the range, while \CPCF{} remains competitive and scales better with the number of trees. 
These experiments reinforce our previous observations: \MaxSAT{} is preferable for hard-voting random forests, but \CPCF{} is most suitable once the model departs from that purely Boolean regime through numerical confidence scores. 

\begin{figure}[t!]
    \centering
    \includegraphics[width=1.0\linewidth]{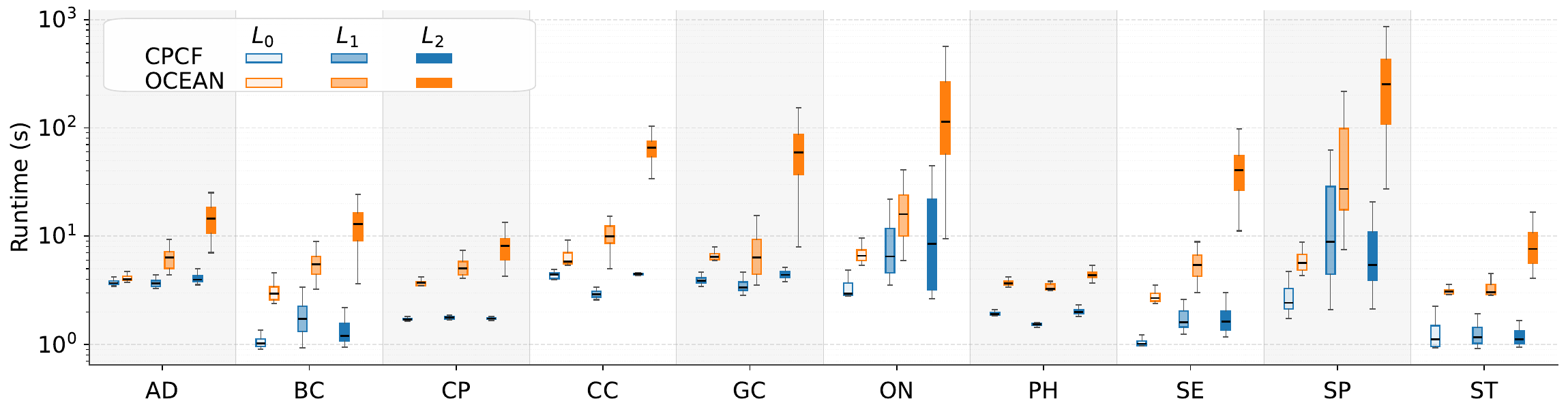}
    \caption{Distribution of total times to optimality for \CPCF\ and \OCEAN\ under \(L_0\), \(L_1\), and \(L_2\)-type cost functions, for the default configuration (100 depth-5 trees) of soft-voting random forests.}
    \label{fig:norms}
\end{figure}

\textbf{Result 4. \CPCF{} remains efficient across different objective choices.}
Figure~\ref{fig:norms} compares the total solution times of \CPCF{} and \OCEAN{} under \(L_0\), \(L_1\), and \(L_2\) on the default ensemble configuration. We observe that \CPCF{} remains faster than \OCEAN{} across all datasets and norms. This gap is especially pronounced under \(L_2\), for datasets with many numerical features. This is coherent with the formulations: in \CPCF{}, changing the norm mostly amounts to changing the coefficients in the objective, in particular through~\eqref{eq:cp_delta_num}. By contrast, the MILP formulation is more sensitive to the objective: under the \(L_2\) norm, it becomes a quadratic program.

\textbf{Result 5. \CPCF{} efficiently accommodates plausibility constraints.}
Table~\ref{tab:isolation-transposed} compares \CPCF{} and \OCEAN{}, the two approaches supporting plausibility constraints, with and without such constraints under the default soft-voting random forest configuration. Plausibility is enforced using an isolation forest of \(100\) trees with a contamination level (proportion of training examples flagged as outliers) of \(10\%\). For each dataset, we report the proportion of plausible explanations among all queries and the median total runtime. Two conclusions emerge. First, the constraint behaves as intended: without it, the proportion of plausible explanations varies substantially across datasets, whereas both formulations return fully plausible counterfactuals once it is enforced. Second, this plausibility layer increases the solution time for both methods, as expected, since it makes the optimization problem more computationally demanding. However, it does not change their relative ranking. Across all the datasets, \CPCF{} remains consistently faster than \OCEAN{}, both with and without the isolation forest. The gap is especially marked on larger or more structured datasets, where the additional plausibility constraints amplify the benefit of stronger global propagation in CP. 

\begin{table}
    \centering 
    \footnotesize
    \setlength{\tabcolsep}{3pt}
    \caption{Effect of isolation forest plausibility constraints on \OCEAN\ and \CPCF\, for the default configuration of soft-voting random forests. For each dataset, \(P\) reports the proportion of plausible explanations among all queries, and \(T\) reports the median total time to optimality in seconds.}
    \label{tab:isolation-transposed}
    \begin{tabular}{llrrrrrrrrrr}
    \toprule
    Method & Metric & AD & BC & CP & CC & GC & ON & PH & SE & SP & ST \\
    \midrule
    \multirow{2}{*}{\textbf{\CPCF-noIF}}
        & \(P\) (\%) & 96 & 22 & 82 & 70 & 68 & 82 & 80 & 18 & 50 & 90 \\
        & \(T\) (s)  & \textbf{4.62} & \textbf{1.87} & \textbf{2.02} & \textbf{3.06} & \textbf{3.36} & \textbf{6.28} & \textbf{2.07} & \textbf{1.62} & \textbf{11.58} & \textbf{1.43} \\
    \midrule
    \multirow{2}{*}{\textbf{\OCEAN-noIF}}
        & \(P\) (\%) & 96 & 26 & 84 & 70 & 72 & 82 & 78 & 18 & 50 & 88 \\
        & \(T\) (s)  & 10.31 & 7.16 & 11.96 & 15.74 & 13.92 & 41.88 & 6.84 & 20.19 & 61.92 & 5.82 \\
    \midrule \midrule
    \multirow{2}{*}{\textbf{\CPCF-IF}}
        & \(P\) (\%) & 100 & 100 & 100 & 100 & 100 & 100 & 100 & 100 & 100 & 100 \\
        & \(T\) (s)  & \textbf{10.92} & \textbf{16.51} & \textbf{17.78} & \textbf{12.85} & \textbf{12.23} & \textbf{21.60} & \textbf{9.29} & \textbf{18.50} & \textbf{43.44} & \textbf{9.51} \\
    \midrule
    \multirow{2}{*}{\textbf{\OCEAN-IF}}
        & \(P\) (\%) & 100 & 100 & 100 & 100 & 100 & 100 & 100 & 100 & 100 & 100 \\
        & \(T\) (s)  & 27.00 & 38.02 & 48.41 & 39.86 & 50.73 & 132.48 & 28.03 & 78.45 & 106.68 & 25.81 \\
    \bottomrule
    \end{tabular}
\end{table}

\section{Conclusion}

We introduced \CPCF, a constraint programming formulation for computing optimal counterfactual explanations in tree ensembles. Across a broad empirical evaluation, \CPCF{} achieves state-of-the-art total time to proven optimality, strong anytime performance, and robust scalability across ensemble size, tree depth, cost functions, and plausibility constraints.
Our results also show that the relative strengths of different mathematical programming paradigms depend on the specific setting.

Future work could extend \CPCF{} in several directions. First, more complex cost functions could be considered, for instance when per-feature costs are interdependent or when feature modifications must comply with a given causal model. Efficient automata representations in CP solvers could, for instance, be leveraged to encode causal dependencies between admissible feature modifications. Second, integrating additional desiderata, such as sparsity or robustness, would naturally lead to multi-objective counterfactual formulations. Providing precise and certifiable characterizations of the trade-offs between these desiderata would be an interesting direction. Finally, adapting \CPCF{} beyond tree ensembles (that are axis-aligned piecewise-constant models) is another promising direction, although it would require moving beyond the current per-feature interval representation.

More broadly, our results suggest that the efficiency of each mathematical programming paradigm depends on how well its modeling primitives match the structure of the problem at hand. This perspective extends beyond counterfactual explanations for tree ensembles to other important problems in trustworthy machine learning, such as verification, robustness analysis, fairness-constrained learning, and sparse explanations, which combine logical, continuous, and combinatorial components. Identifying which paradigm best captures these structures, or how to combine paradigms when they coexist, is an important direction for future work.

\bibliographystyle{plainnat}
\bibliography{references}

\appendix
\newpage
\section{Details on the Construction of Figure~\ref{fig:decision-diagram}}
\label{app:decision-diagram}

Figure~\ref{fig:decision-diagram} is obtained by fitting a small decision-diagram meta-classifier to our experimental benchmark results, using the optimal decision-diagram learning method introduced by~\citet{florio2022optimaldecisiondiagramsclassification} with a \(600\) second time limit. Each training row corresponds to one configuration--metric pair and is described by simple dataset and ensemble descriptors: the numbers of numerical, ordinal, categorical, and binary features; the ensemble type and voting regime; the number of estimators; the maximum depth; the total, per-tree mean, and per-tree maximum numbers of tree nodes; the total, mean, and maximum numbers of split levels associated with each feature; the objective norm; whether isolation forest plausibility is enabled; and the performance metric of interest, either median total running time or median solver-only time, for which model-building time is ignored. The resulting training dataset contains \(3{,}380\) rows, corresponding to \(2 \times 1{,}690\): each of the \(1{,}690\) unique experimental configurations contributes one row for median total time and one row for median solver-only time. Each row is labeled with the best-performing method among \MACE{}, \OCEAN{}, \MaxSAT{}, and \CPCF{} for the corresponding configuration and metric.

\MACE{} is the best-performing method on only one row in the full \(3{,}380\)-row table. Since the purpose of the decision diagram is to summarize the practically relevant regimes rather than isolate a single outlier configuration, we remove that row and train the final diagram on the remaining \(3{,}379\) rows using only the three dominant exact paradigms: \CPCF{}, \OCEAN{}, and \MaxSAT{}. The percentages displayed in the nodes report the empirical fraction of retained rows routed through each node. The resulting decision diagram (\ref{fig:decision-diagram}) achieves an accuracy of \(87{.}79\%\) on the full dataset.

\section{\MaxSAT{} Formulation and Soft-Voting Extension}
\label{app:maxsat-soft-voting}

\citet{inbookMaxSAT} formulate hard-voting random forest counterfactual search as a partial weighted \MaxSAT{} instance. Using our notation, the formulation is naturally expressed through a set of hard clauses \(\Phi_{\mathrm{hard}}\) and a set of weighted soft clauses \(\Phi_{L_1}\). Note that in a partial weighted \MaxSAT{} instance, hard clauses must be satisfied, whereas soft clauses may be violated at a cost. Each soft clause is associated with a nonnegative weight, and the objective is to minimize the total weight of violated soft clauses. Here, the soft clauses encode the distance from the original point \(\hat{\mathbf{x}}\): preserving the original value of a feature satisfies the corresponding soft clauses, while moving away from it violates clauses whose weights add up to the \(L_1\) cost of that move.

\paragraph{Variables.}
For each tree \(t \in \mathcal{T}\) and leaf \(\ell \in \mathcal{L}_t\), let \(z_{t,\ell}\) indicate that the counterfactual reaches leaf \(\ell\). For each categorical feature \(f \in \mathcal{F}_C\) and category \(j \in \{1, \ldots, k_f\}\), let \(\nu_{f,j}\) be the corresponding one-hot literal. Binary features $\mathcal{F}_B$ are special cases of categorical features and can be represented with a single Boolean literal; in what follows, we treat them together with categorical features. For each numerical or ordinal feature \(f \in \mathcal{F}_N \cup \mathcal{F}_O\) and threshold \(\tau_{f,m}\), let \(t_{f,m}\) (such that $m \in \{1, \ldots, k_f\}$) denote the literal stating that the counterfactual lies on the left side of \(\tau_{f,m}\).

\paragraph{Hard clauses.}
Feature validity is encoded by
\[
\Phi_{\mathrm{dom}}
=
\Phi_{\mathrm{cat}} \land \Phi_{\mathrm{thr}},
\]
where
\[
\Phi_{\mathrm{cat}}
=
\bigwedge_{f \in \mathcal{F}_C}
\Bigg(
\bigvee_{j=1}^{k_f} \nu_{f,j}
\land
\bigwedge_{\substack{1 \le j < j' \le k_f}}
(\neg \nu_{f,j} \lor \neg \nu_{f,j'})
\Bigg),
\]
which means that for all categorical features, exactly one category is selected, and
\[
\Phi_{\mathrm{thr}}
=
\bigwedge_{f \in \mathcal{F}_N \cup \mathcal{F}_O}
\bigwedge_{m=1}^{k_f-1}
(\neg t_{f,m} \lor t_{f,m+1}),
\]
which enforces consistency among threshold literals: crossing a threshold implies crossing all smaller thresholds, or equivalently, lying on the left side of a threshold implies lying on the left side of all larger thresholds.

For each \(t \in \mathcal{T}\), exactly one leaf must be active:
\[
\Phi_{\mathrm{leaf}}
=
\bigwedge_{t \in \mathcal{T}}
\left(
\bigvee_{\ell \in \mathcal{L}_t} z_{t,\ell}
\;\land\;
\bigwedge_{\substack{\ell,\ell' \in \mathcal{L}_t\\ \ell \neq \ell'}}
(\neg z_{t,\ell} \lor \neg z_{t,\ell'})
\right).
\]

Let \(L(t,\ell)\) denote the set of literals induced by the root-to-leaf path of leaf \(\ell\), namely literals of the form \(t_{f,m}\) or \(\neg t_{f,m}\) for numerical and ordinal splits, and \(\nu_{f,j}\) or \(\neg \nu_{f,j}\) for categorical splits. Path consistency is then encoded by
\[
\Phi_{\mathrm{path}}
=
\bigwedge_{t \in \mathcal{T}}
\bigwedge_{\ell \in \mathcal{L}_t}
\bigwedge_{\lambda \in L(t,\ell)}
(\neg z_{t,\ell} \lor \lambda).
\]

If \(y(\ell)\) is the class label of leaf \(\ell\), the hard-voting target-class condition is encoded by
\[
\Phi_{\mathrm{class}}
=
\bigwedge_{y \in \mathcal{Y}\setminus\{y^\star\}}
\mathrm{Card}\!\left(
\sum_{t \in \mathcal{T}} \sum_{\ell \in \mathcal{L}_t:\, y(\ell)=y^\star} z_{t,\ell}
\ge
\sum_{t \in \mathcal{T}} \sum_{\ell \in \mathcal{L}_t:\, y(\ell)=y} z_{t,\ell}
+\varepsilon_c
\right).
\]
The full hard part is therefore
\[
\Phi_{\mathrm{hard}}
=
\Phi_{\mathrm{dom}}
\land
\Phi_{\mathrm{leaf}}
\land
\Phi_{\mathrm{path}}
\land
\Phi_{\mathrm{class}}.
\]

\paragraph{\(L_1\) soft clauses.}
To facilitate comparison with the other formulations, we retain only the \(L_1\) part of the original objective. For a binary or categorical feature \(f\), let \(\rho_f\) be the literal preserving the current value of the query \(\hat{\mathbf{x}}\). The corresponding soft clauses are
\[
\Phi_{L_1}^{\mathrm{cat}}
=
\bigwedge_{f \in \mathcal{F}_B \cup \mathcal{F}_C}
(\rho_f,\; wt=\alpha_f).
\]
where \(\alpha_f > 0\) denotes the weight assigned to feature \(f\) in the weighted \(L_1\) objective.
For a numerical feature \(f\), let \(\hat{x}_f \in (\tau_{f,j},\tau_{f,j+1}]\), and define
\[
\bar{\tau}_{f,m}
=
\frac{\tau_{f,m}-\mathrm{lb}(f)}{\mathrm{ub}(f)-\mathrm{lb}(f)},
\qquad
\bar{x}_f
=
\frac{\hat{x}_f-\mathrm{lb}(f)}{\mathrm{ub}(f)-\mathrm{lb}(f)}.
\]
The exact incremental \(L_1\) encoding is
\[
\Phi_{L_1}^{\mathrm{num}}(f)
=
\bigwedge_{k=1}^{j-1}
(\neg t_{f,k},\; wt=\alpha_f(\bar{\tau}_{f,k+1}-\bar{\tau}_{f,k}))
\]
\[
\land
(\neg t_{f,j},\; wt=\alpha_f(\bar{x}_f-\bar{\tau}_{f,j}))
\land
(t_{f,j+1},\; wt=\alpha_f(\bar{\tau}_{f,j+1}-\bar{x}_f))
\]
\[
\land
\bigwedge_{k=j+2}^{k_f}
(t_{f,k},\; wt=\alpha_f(\bar{\tau}_{f,k}-\bar{\tau}_{f,k-1})).
\]
This encoding exploits the monotonicity of the threshold literals. Since \(t_{f,m}\) denotes the condition \(x_f \le \tau_{f,m}\), the original value \(\hat{x}_f \in (\tau_{f,j},\tau_{f,j+1}]\) satisfies \(\neg t_{f,k}\) for all thresholds below \(\hat{x}_f\), and \(t_{f,k}\) for all thresholds above \(\hat{x}_f\). The soft clauses therefore reward keeping the counterfactual on the same side of each threshold as \(\hat{x}_f\). If the counterfactual crosses a threshold, the corresponding soft clause is violated and its weight contributes to the objective. The weights are chosen as normalized interval lengths, so the sum of violated weights is exactly the weighted normalized \(L_1\) displacement along feature \(f\).
For ordinal features, the same construction is used after replacing each threshold by the nearest admissible value on the appropriate side. The full soft-clause family is then
\[
\Phi_{L_1}
=
\Phi_{L_1}^{\mathrm{cat}}
\land
\bigwedge_{f \in \mathcal{F}_N \cup \mathcal{F}_O}
\Phi_{L_1}^{\mathrm{num}}(f).
\]

\paragraph{Soft-voting extension.}
The original formulation is specific to hard-voting tree ensembles. In our benchmark, we extend it to support soft-voting tree ensembles. We keep \(\Phi_{\mathrm{dom}}\), \(\Phi_{\mathrm{leaf}}\), and \(\Phi_{\mathrm{path}}\) unchanged, and replace \(\Phi_{\mathrm{class}}\) by
\[
\Phi_{\mathrm{class}}^{\mathrm{soft}}
=
\bigwedge_{y \in \mathcal{Y}\setminus\{y^\star\}}
\mathrm{PB}\!\left(
\sum_{t \in \mathcal{T}} \sum_{\ell \in \mathcal{L}_t}
\bar{p}_{t,\ell}^{(y^\star,y)} z_{t,\ell}
\ge
\varepsilon_c
\right),
\]
where \(\bar{p}_{t,\ell}^{(y^\star,y)}\) is an integer scaling of \(p_{t,\ell,y^\star}-p_{t,\ell,y}\), and \(\mathrm{PB}(\cdot)\) denotes a CNF encoding of the corresponding pseudo-Boolean inequality~\citep{pblib.sat2015, pseudobool}.

\section{Integrating Isolation Forest Plausibility Constraints into \CPCF{}}
\label{app:isolation-details}

For each target class \(y^\star\), we train a separate isolation forest on the training examples labeled \(y^\star\). This model estimates the training-data distribution within the target class, yielding a plausibility criterion aligned with the target region rather than with the full data distribution.

Let \(\mathcal{T}_{\mathrm{IF}}^{(y^\star)}\) denote the isolation forest associated with target class \(y^\star\). As in the explained tree ensemble, each isolation tree contributes exactly one active leaf, and each active leaf is assigned a corrected path length. The path of the counterfactual through the isolation forest is encoded in the same way as for the explained tree ensemble, using leaf-selection variables and path-consistency constraints~\eqref{eq:leaf}--\eqref{eq:cp_binary_path}.

For a leaf \(\ell\) of an isolation tree \(t \in \mathcal{T}_{\mathrm{IF}}^{(y^\star)}\), let \(\mathrm{depth}_t(\ell)\) denote its depth, i.e., the length of the path from the root of \(t\) to \(\ell\), and let \(m_{t,\ell}\) be the number of training samples that reached \(\ell\) during isolation forest construction. The corrected path length is then defined as
\begin{equation*}
L_{t,\ell}=
\begin{cases}
\mathrm{depth}_t(\ell), & \text{if } m_{t,\ell}\le 1,\\[1mm]
\mathrm{depth}_t(\ell)+c(m_{t,\ell}), & \text{if } m_{t,\ell}>1,
\end{cases}
\label{eq:if_path_length}
\end{equation*}
where \(c(m)\) is the standard isolation forest correction term,
\begin{equation*}
c(m)=
\begin{cases}
0, & \text{if } m\le 1,\\[1mm]
1, & \text{if } m=2,\\[1mm]
2\ln(m-1)+2\gamma-\dfrac{2(m-1)}{m}, & \text{if } m>2,
\end{cases}
\label{eq:if_correction}
\end{equation*}
and \(\gamma \approx 0.57721\) is the Euler--Mascheroni constant.

Let \(z_{t,\ell} \in \{0,1\}\) denote the leaf-selection variable for leaf \(\ell\) in isolation tree \(t\). The average corrected path length of the counterfactual through the target-class isolation forest is then computed with:
\begin{equation}
H(\mathbf{x})=
\frac{1}{|\mathcal{T}_{\mathrm{IF}}^{(y^\star)}|}
\sum_{t \in \mathcal{T}_{\mathrm{IF}}^{(y^\star)}}
\sum_{\ell \in \mathcal{L}_t^{\mathrm{IF}}}
L_{t,\ell}\,z_{t,\ell}.
\label{eq:if_avg_length}
\end{equation}

The corresponding anomaly score is
\[
\mathrm{score}(\mathbf{x})=-2^{-H(\mathbf{x})/c_{\max}},
\]
where \(m_{\max}\) is the isolation forest subsample size and \(c_{\max}=c(m_{\max})\). The isolation forest decision function is then defined as
\[
\mathrm{decision}(\mathbf{x})
=
\mathrm{score}(\mathbf{x})-\mathrm{offset},
\]
where \(\mathrm{offset}\) is a constant chosen from the fitted isolation forest to match the prescribed \emph{contamination level}, namely the proportion of training examples classified as outliers. A counterfactual is considered plausible if and only if
\[
\mathrm{decision}(\mathbf{x}) \ge 0.
\]

Because the exponential form cannot be represented directly in the CP model, we reformulate this condition as an equivalent linear lower bound on the average path length. The final plausibility constraint is therefore enforced as:
\begin{equation}
H(\mathbf{x}) \ge H_{\min},
\label{eq:if_final_constraint}
\end{equation}
where $H_{\min}=-c_{\max}\log_2(-\mathrm{offset})$. Intuitively, examples isolated early or falling into leaves with few training samples have shorter corrected path lengths. They therefore have lower values of \(H(\mathbf{x})\) and are more likely to be classified as outliers by the isolation forest.

Both \CPCF{} and \OCEAN{} use this corrected path-length formulation. This slightly refines the original \OCEAN{} isolation forest constraint, which was based on the selected leaf depths and a fixed average-depth threshold. In Table~\ref{tab:isolation-transposed}, the ``noIF'' variants do not enforce Equation~\eqref{eq:if_final_constraint}; their plausibility rates are therefore computed \emph{a posteriori} using the same target-class-specific isolation model. In contrast, the ``IF'' variants enforce this constraint during optimization, and therefore produce \(100\%\) plausible counterfactuals by construction.

\section{Implementation Details}
\label{app:backend-details}

\subsection{Numerical Confidence Scores Handling}

The mathematical programming formulations considered in this work all solve the same optimization problem. However, while \OCEAN{} can model confidence scores using continuous variables, as discussed in Appendix~\ref{app:ocean-details}, the other paradigms require a discrete numerical representation.

In \CPCF{}, confidence scores used to enforce the target-class prediction are represented as integers. We therefore multiply these scores by \(10^9\) and round them to the nearest integer. In soft-voting \MaxSAT{}, pairwise score differences are similarly scaled before the pseudo-Boolean encoding, so that the target-class constraints can be represented in CNF form. For \MACE{}, we retain the numerical treatment of the authors' implementation: the random forest encoding uses real-valued auxiliary variables for tree-level class probabilities and enforces the forest prediction by comparing the corresponding sums.

These transformations do not change the intended target-class comparison in Equation~\eqref{eq:target_class}; they only adapt its numerical representation to the requirements of the underlying solver. In \CPCF{} and soft-voting \MaxSAT{}, confidence scores are converted to integer quantities by scaling them by \(10^9\) and rounding. This introduces at most \(5\times 10^{-10}\) absolute error per scaled score, which is below the precision at which the confidence scores are exposed by the tree-ensemble implementations and below the numerical thresholds used in our experiments. Thus, the discrete encodings preserve the same effective target-class comparisons as the original floating-point scores.

For consistency, we also tighten the numerical tolerances of the MILP solver used by \OCEAN{}. This avoids comparing \OCEAN{} under a looser default feasibility or integrality tolerance than the discrete encodings used by \CPCF{} and \MaxSAT{}. Consequently, all formulations are evaluated under the same effective precision regime, and observed differences in performance are not attributable to different numerical tolerances across solver backends.

\subsection{\OCEAN-Specific Implementation Details}
\label{app:ocean-details}

To align the numerical precision of \OCEAN{} with that of the other baselines, we make a few adaptations to the implementation. In particular, we set the \texttt{Gurobi} parameters \texttt{FeasibilityTol=1e-9}, \texttt{IntFeasTol=1e-9}, and \texttt{IntegralityFocus=1}, and reduce the prediction margin from \(10^{-4}\) to \(10^{-7}\), matching the value used in \CPCF{}. These changes do not modify the formulation itself, but make the returned counterfactuals more faithful to the reference tree-ensemble predictions.

A second difference concerns the target class. While \citet{parmentier2021optimal} report experiments in which counterfactuals are generated only toward target class \(1\), we generate counterfactual explanations for any admissible target class \(y^\star \in \mathcal{Y}\setminus\{h_{\mathcal{T}}(\hat{\mathbf{x}})\}\). This choice is consistent with the multiclass setting considered in our benchmark and with the unified problem definition used throughout the paper.

\section{Additional Experimental Details}
\label{app:experimental-details}

This appendix provides additional details on the experimental protocol used in Section~\ref{sec:experimental-results}.
\subsection{Model Training and Query Generation}

For each configuration, dataset, and seed, we train the ensemble on the full processed dataset and then sample \(50\) query instances without replacement from the same dataset. For binary tasks, the target class is set to the opposite label. For multiclass tasks, the target class is selected uniformly at random from \(\mathcal{Y}\setminus\{h_{\mathcal{T}}(\hat{\mathbf{x}})\}\).

Soft-voting random forests use the class scores returned by the fitted \texttt{scikit-learn} leaves. Hard-voting random forests are obtained from the same trained forests by replacing each leaf score vector with a one-hot vote for its majority class, while keeping the tree structure unchanged. XGBoost is evaluated only in the soft-voting regime.

For the eight datasets inherited from \citet{parmentier2021optimal}, we keep the same preprocessing pipeline, but do not enforce the dataset-specific actionability restrictions used in that paper. This choice avoids introducing dataset-dependent feasibility constraints that could interact differently with the competing formulations. Instead, all methods are evaluated on the same unrestricted processed feature space, corresponding to the largest common search space induced by the preprocessing.

\subsection{Evaluation and Plotting Conventions}

All reported runtimes are total times, defined as model-construction time plus solving time, unless explicitly stated otherwise. Indeed, before an optimization formulation can be solved, the corresponding model must first be instantiated through the solver API: variables must be created, domains specified, and constraints posted. This model-construction phase, which we refer to as build time, can be non-negligible and may vary substantially across optimization paradigms and solver interfaces. Reporting total time, therefore, captures the end-to-end computational cost incurred when solving a query.
In the scaling plots, runs that do not prove optimality within the time limit are all censored at \(900\) seconds before computing medians.

The anytime curves compare only \CPCF{} and \OCEAN{}, since these are the only two approaches in our benchmark that expose intermediate incumbents. They are restricted to instances solved to optimality by both methods, which covers a wide majority of the queries (more than $99$\% of the queries in the worst case in our experiments). For each such instance, the normalized error at time \(t\) is defined as
\[
\frac{\mathrm{cost}_t-\mathrm{cost}^\star}{\mathrm{cost}_{\max}-\mathrm{cost}^\star},
\]
where \(\mathrm{cost}_t\) is the best incumbent objective available at time \(t\), \(\mathrm{cost}^\star\) is the final optimal objective, and \(\mathrm{cost}_{\max}\) is the worst objective value observed across the two incumbent traces for that instance.

\end{document}